\documentclass{article}

\usepackage{arxiv}

\usepackage[utf8]{inputenc} % allow utf-8 input
\usepackage[T1]{fontenc}    % use 8-bit T1 fonts
\usepackage{hyperref}       % hyperlinks
\usepackage{url}            % simple URL typesetting
\usepackage{booktabs}       % professional-quality tables
\usepackage{amsfonts}       % blackboard math symbols
\usepackage{nicefrac}       % compact symbols for 1/2, etc.
\usepackage{microtype}      % microtypography
\usepackage{lipsum}
\usepackage{graphicx}

\usepackage[inline]{enumitem}
\usepackage{cite}
\usepackage{natbib}
\graphicspath{ {./images/} }

\title{The Large Labelled Logo Dataset (L3D): A Multipurpose and Hand-Labelled Continuously Growing Dataset}

\author{
 Asier Gutiérrez-Fandiño\\
  Barcelona Supercomputing Center\\
  \texttt{asier.gutierrez@bsc.es} \\
   \And
 David Pérez-Fernández\\
  Universidad Autónoma de Madrid\\
  \texttt{david.perez@inv.uam.es} \\
  \And
 Jordi Armengol-Estapé\\
  Barcelona Supercomputing Center\\
  \texttt{jordi.armengol@bsc.es} \\
}

\begin{document}
\maketitle
\begin{abstract}
In this work, we present the Large Labelled Logo Dataset (L3D), a multipurpose, hand-labelled, continuously growing dataset. It is composed of around 770k of color 256x256 RGB images extracted from the European Union Intellectual Property Office (EUIPO) open registry. Each of them is associated to multiple labels that classify the figurative and textual elements that appear in the images. These annotations have been classified by the EUIPO evaluators using the Vienna classification, a hierarchical classification of figurative marks. We suggest two direct applications of this dataset, namely, logo classification and logo generation.
\end{abstract}

\section{Introduction}

Having large labelled datasets is of the uttermost importance for building deep learning applications. The importance of annotated datasets for the advancement of machine learning is unquestionable. Annotated datasets are needed to understand the complexity of human perception and knowledge acquisition. Thanks to the efforts of many researchers, Deep Learning has large and varied datasets. Datasets such as MNIST or CIFAR are at the basis of definitive advances in automated image recognition. It is important that these datasets represent real cases, their contents are sufficiently refined and the amount of data is representative in variety and quantity of data.

The data set collected corresponds to 0.7M images obtained from the EUIPO trademark register. When a company or individual registers in the EUIPO a logo associated with a company, product or service, the registration is made by standardizing the image used and associating multiple labels describing the figurative elements contained in the image. The objective of the EUIPO evaluators is to verify that no new registration of a logo already exists in the image database nor any of its elements. 

To do so, the human staff use multiple tags for each image following the Vienna Classification:
\url{https://www.wipo.int/classifications/vienna/en/}
The Vienna Classification of Classification (VCL), whose latest version 9, which is mandatory as of January 1, 2023, can be viewed here:
\url{https://www.wipo.int/classifications/nivilo/vienna.htm}
is composed of 29 categories organized in a structure of 3 hierarchical levels.

\section{Related Work}
There is a wide variety of large image datasets suitable for Deep Learning. The well-known MNIST \citep{lecun-mnisthandwrittendigit-2010}, CIFAR \citep{Krizhevsky09learningmultiple} and Imagenet \citep{deng2009imagenet} are on the hall of fame and are used as de facto sources for validating new Deep Learning architectures. These datasets started to be used for image classification tasks. However, as Deep Learning architectures unlocked additional capabilities, these datasets have been used for other tasks. At the time this was happening new datasets appeared.

Human perception is based on the understanding of objects and scenes. Multitude of applications, from the development of autonomous cars \citep{Argoverse, yu2020bdd100k, Cordts2016Cityscapes, 1812.05752}, digitizing text using optical character recognition (OCR) systems, improving video game graphics \citep{Richter_2021}, transforming drawings to images \citep{park2019semantic} to transferring drawing style \citep{li2018learning}.

The most recent datasets and, in turn, most closely linked to human perception are those of DALL-E \citep{ramesh2021zeroshot} and those being developed by EleutherAI. Both datasets, from the network images and their descriptions, train a text-to-image model.

The dataset presented in this article consists of digital figurative images created by humans, projecting shapes and objects, according to the human perception of them. In many cases it is an artistic or synthetic view of the figurative elements that compose it. In addition, this data set is rigorously labeled with multiple tags that have a very detailed granularity organized in a hierarchical fashion. Since images may contain text, it also includes the text that is present in the image.

A very interesting work on how application of supervised learning algorithms in order to automate the manual process of labeling trademark images with Vienna codes can be found in \cite{Uzairi1606039}. This work compares different deep learning algorithms, namely, CNN, recurrent neural networks (LSTMs and GRUs), Support Vector Machines, Decision Trees, Random Forests and Naive Bayes models.

\section{L3D Dataset}
\subsection{General dataset description}
The European Union Intellectual Property Office (EUIPO), founded in 1994, is the European Union Agency  responsible for the registration of the European Union trade mark and the registered Community design, two unitary intellectual property rights valid across the 27 Member States of the EU. The EUIPO stores all the information from trademarks filled at the European Union and distributes them in order to avoid the creation of similar trademarks.

The EUIPO creates monthly incremental versions of the current year and once the year has passed, creates a final version of that year. In our work we selected images until 2020 for simplicity. However, we encourage researchers parameterizing the download to obtain next years.

EUIPO’s Open Data Platform is a tool to make all the trade mark and design information available and transparent to its users, with quick and efficient access. A new application makes the Register available to the general public for bulk download.
\footnote{\url{https://euipo.europa.eu/ohimportal/en/open-data}}

EUIPO started in 1998 making its trade mark database available online to firms and national Industrial Property offices under a license contract. Now EUIPO is opening their databases to all users, free of charge, updated daily and without requiring a license.\footnote{\url{https://euipo.europa.eu/ohimportal/en/news?p_p_id=csnews_WAR_csnewsportlet&p_p_lifecycle=0&p_p_state=normal&p_p_mode=view&p_p_col_id=column-1&p_p_col_count=2&journalId=3584337&journalRelatedId=manual/}}

In order to browse the contents of the database in a user-friendly way, the EUIPO has a trademark viewer web-client. \footnote{\url{https://www.tmdn.org/tmview/}}

\subsection{Vienna classification}
Images Contained in this database are classified using the Vienna Classification (International Classification of the Figurative Elements of Marks)\footnote{\url{https://www.wipo.int/classifications/vienna/en/}}. Vienna Classification (VCL) is an international classification system used to classify the figurative elements of logos or images registered by companies (company, product or services registered trademarks).

The Vienna Classification of figurative elements was agreed by the World Intellectual Property Organization (WIPO) in the so-called Vienna Agreement on June 12, 1973, and came into force on August 9, 1985. The use of the Vienna Classification by national industrial property offices is intended to facilitate the registration of trademarks containing figurative elements by coding them according to a single classification system at the international level. This facilitates the search for similar images of other registered trademarks and avoids having to reclassify images when they are registered in multiple industrial property offices at the international level.

The International Bureau of WIPO also applies the Vienna Classification in the framework of the Madrid System for the international registration of marks. Around 60 offices in the world apply the Vienna Classification. In addition, three regional organizations, namely the African Regional Intellectual Property Organization (ARIPO), the Benelux Organisation for Intellectual Property (BOIP) and the European Union Intellectual Property Office (EUIPO), use the Vienna Classification.

The VCL is a hierarchical system classifying all figurative elements into 29 categories, further divided into divisions and sections. Additional information can be found at WIPO VCL site \footnote{\url{https://www.wipo.int/classifications/vienna/en/}}. Table \ref{tab:tab} shows the categories of the VCL.

\begin{table}[]
\centering
\begin{tabular}{|c|l|}
\hline
\textbf{Category} & \textbf{Description}                                            \\ \hline
1        & Celestial Bodies, Natural Phenomena, Geographical Maps          \\ \hline
2        & Human Beings                                                    \\ \hline
3        & Animals                                                         \\ \hline
4        & Supernatural, Fabulous, Fantastic or Unidentifiable Beings      \\ \hline
5        & Plants                                                          \\ \hline
6        & Landscapes                                                      \\ \hline
7        & Constructions, Structures for Advertisements, Gates or Barriers \\ \hline
8        & Foodstuffs                                                      \\ \hline
9        & Textiles, Clothing, Sewing Accessories, Headwear, Footwear      \\ \hline
10        & Tobacco, Smokers’ Requisites, Matches, Travel Goods, Fans, Toilet Articles      \\ \hline
11        & Household Utensils      \\ \hline
12        & Furniture, Sanitary Installations      \\ \hline
13        & Lighting, Wireless Valves, Heating, Cooking or Refrigerating Equipment,\\
        &  Washing Machines, Drying Equipment      \\ \hline
14        & Ironmongery, Tools, Ladders      \\ \hline
15        & Machinery, Motors, Engines      \\ \hline
16        & Telecommunications, Sound Recording or Reproduction, Computers, \\
        &  Photography, Cinematography, Optics      \\ \hline
17        & Horological Instruments, Jewelry, Weights and Measures      \\ \hline
18        & Transport, Equipment for Animals      \\ \hline
19        & Containers and Packing, Representations of Miscellaneous Products      \\ \hline
20        & Writing, Drawing or Painting Materials, Office Requisites, Stationery and \\
        &  Booksellers’ Goods      \\ \hline
21        & Games, Toys, Sporting Articles, Roundabouts      \\ \hline
22        & Musical Instruments and their Accessories, Music Accessories, Bells, \\
        &  Pictures, Sculptures      \\ \hline
23        & Arms, Ammunition, Armour      \\ \hline
24        & Heraldry, Coins, Emblems, Symbols      \\ \hline
25        & Ornamental motifs, Surfaces or backgrounds with ornaments      \\ \hline
26        & Geometrical figures and Solids      \\ \hline
27        & Forms of writing, Numerals      \\ \hline
28        & Inscriptions in various characters      \\ \hline
29        & Color      \\ \hline
98        & Other Marks      \\ \hline
99        & Too bad logo      \\ \hline
\end{tabular}
\label{tab:tab}
\caption{Level 1 classification of the Vienna code convention.}
\end{table}

\subsection{Dataset Download \& Preparation}
We make freely available the scripts to download and build the dataset\footnote{\url{https://github.com/lhf-labs/tm-dataset}} as well as the link to download it directly.\footnote{\url{https://doi.org/10.5281/zenodo.5771006}} Each script is modular and is supposed to be adapted to the different needs researchers might have. The pipeline proceeds as follows.

\begin{enumerate}
    \item Download: connect to the EUIPO's FTP server and download the files.
    \item Process: previously downloaded files are unziped. Extracted images are renamed as UUIDv4 and their metadata is written into a file. Metadata includes the text contained in the image, the label associated to it and the date of the trademark.
    \item Statistics: compute statistics to obtain number of images, minimum image size, maximum image size, standard deviation of image size and mean image size.
    \item Filter: remove invalid images (i.e. corrupted images) and too small images whose width or height is less than 20 pixels.
    \item Fix (image): transform the TIFF to JPG, rename the images to JPG and set images in fixed sizes of $256 \times 256$ and rename.
    \item Fix (metadata): remove metadata of the invalid images.
    \item Clean: Remove invalid images.
\end{enumerate}

The dataset will consist of $256$ images. Note that the total storage requirements are more than $256$ GB as of December 2021.

\subsection{Statistics}

\begin{figure}[t]
\includegraphics[scale=0.8]{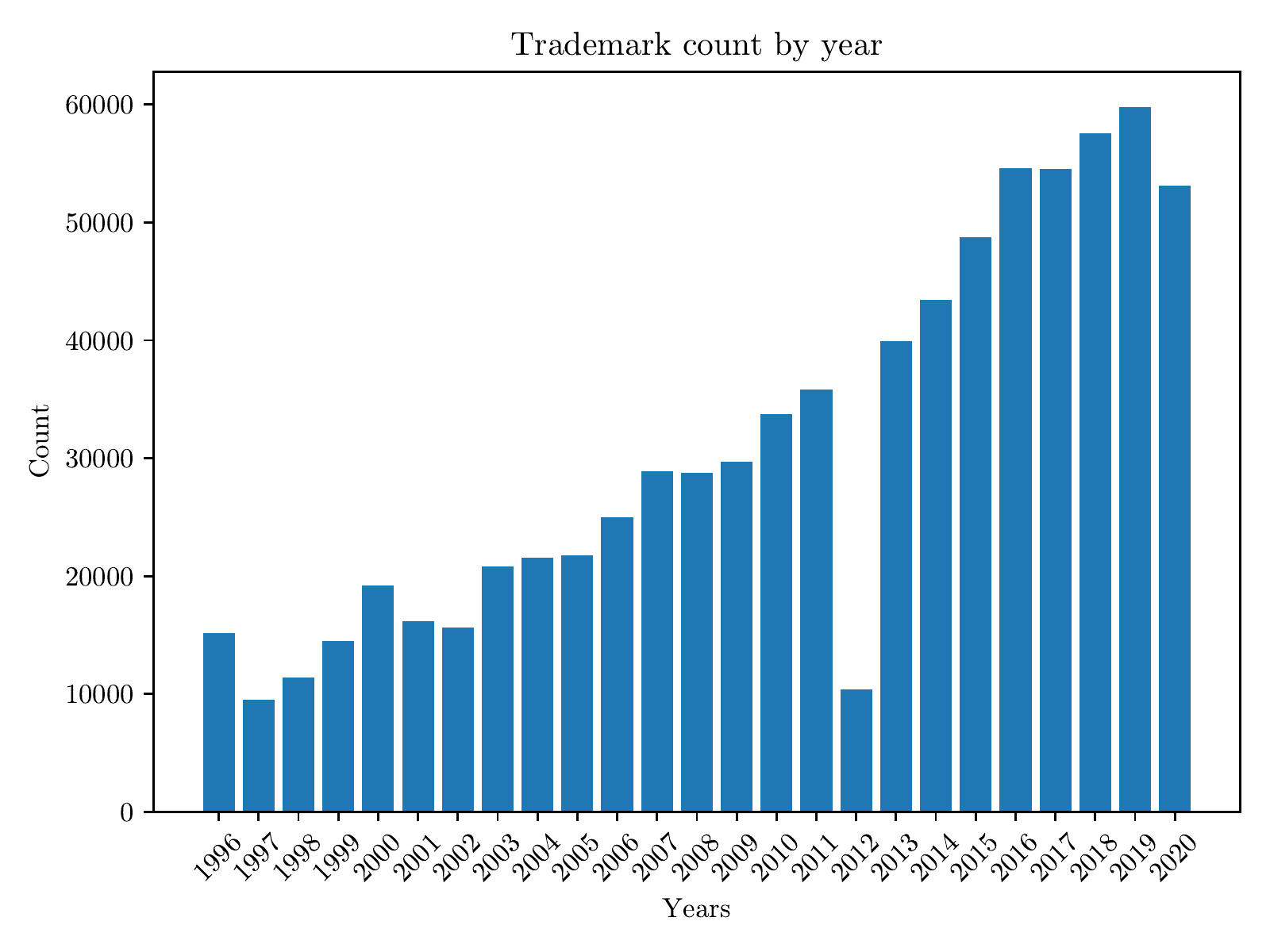}
\label{fig:stats}
\centering
\end{figure}

\begin{figure}[t]
\includegraphics[scale=0.8]{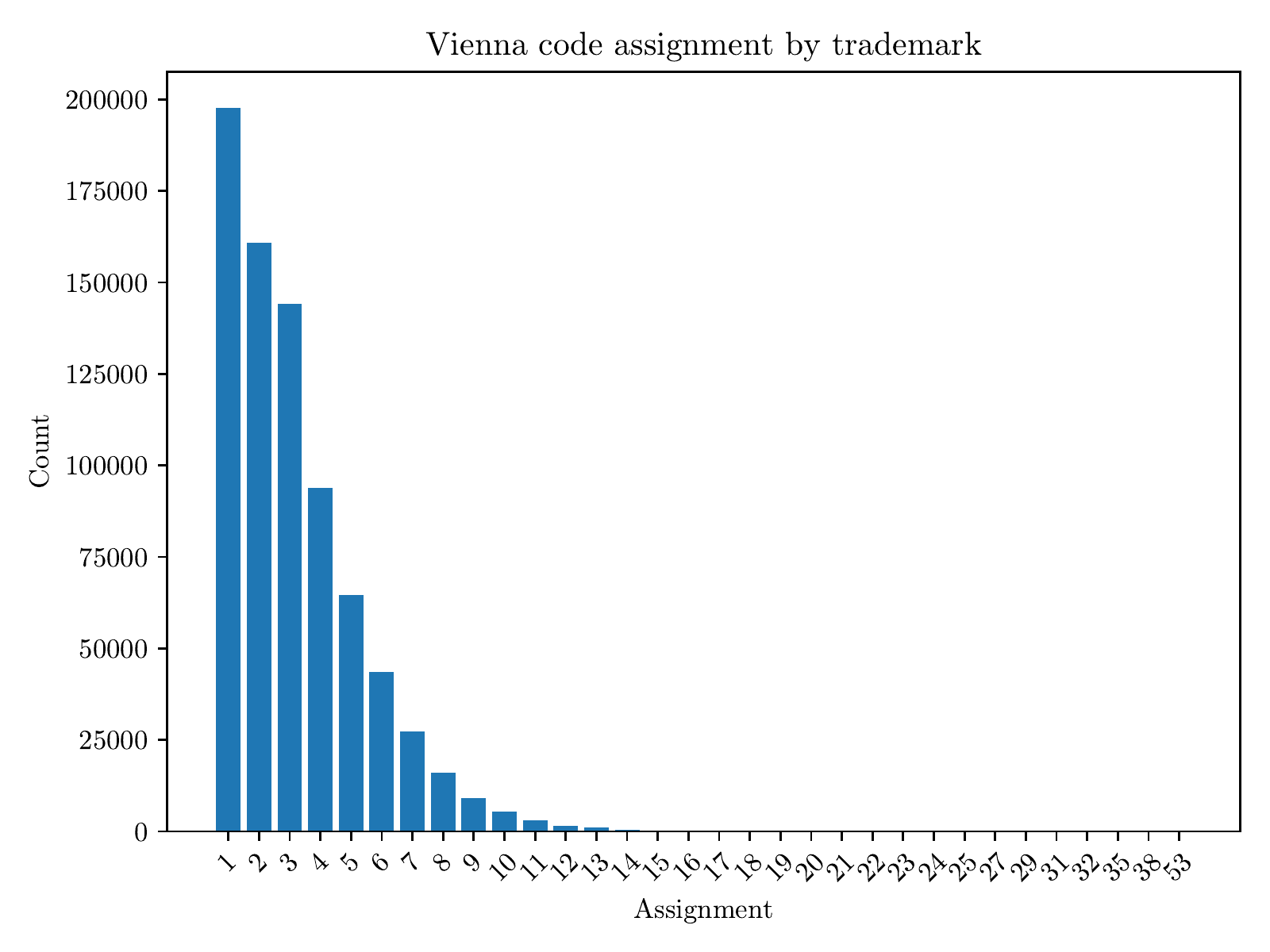}
\label{fig:stats2}
\centering
\end{figure}

Figure \ref{fig:stats} shows the trademark count by year, while Figure \ref{fig:stats2} shows the trademark assignment by category. The dataset is clearly imbalanced.

%[frecuencia de aparición de las etiquetas. Que quede %claro que NO está bien balanceado]
%[histograma número de multietiquetas por imagen]
%[elementos especiales: 
%Category	27	FORMS OF WRITING, NUMERALS
%Category	28	INSCRIPTIONS IN VARIOUS CHARACTERS
%Category	29	COLOURS]

\subsection{Labelling}
%clasificación de Viena
The Vienna Classification is an international classification system used to classify the figurative elements of marks. The complete title of the Classification is International Classification of the Figurative Elements of Marks.

%[1. Clasif Viena ... continuar]

%[2. importancia dataset: multilabel, jerárquico, multiples objetos identificados en la misma imagen, se identifican elementos textuales]

%[3. Principales clases]

\section{Applications}

We foresee several potential applications of this dataset:
\begin{enumerate}
    \item Unconditional trademark generation.
    \item Conditional trademark generation.
    \item Multi-label logo classification (Vienna classification).
    \item Optical Character Recognition.
    \item Conditional trade
    \item Image segmentation.
    \item Image retrieval.
\end{enumerate}
%Aparte de las tareas presentadas en los baselines, proponemos varias tareas adicionales que creemos pueden servir para el benchmarking de nuevos sistemas.

%Dado que, tal y como se comenta en la descripción del dataset, se tiene el texto contenido en el logo, proponemos una tarea de Optical Character Recognition (OCR). Se tiene la imágen y el texto que esta contiene, el objetivo es que el sistema prediga el texto contenido en la imagen.

%Esta tarea de OCR que se propone es extremadamente difícil ya que, como se ha mencionado en el apartado de descripción de dataset, la codificación de los textos en los logos se lleva a cabo de una manera artística. Por tanto, un sistema que pueda ser bueno en esta tarea, significa que está interiorizando la capacidad de interpretación artística de un humano.

%Constrained trademark generation (both labels and text)

%Image segmentation

%Image retrieval %\footnote{\url{https://www.tmdn.org/tmview/}}

\section{Baselines}

In this section, we describe the baselines we built for two of the applications we mentioned, namely, logo generation and multi-label logo classification.

\subsection{Trademark Generation}

\textbf{Awaiting results}

\subsection{Multi-Label Classification}
%\url{https://www3.wipo.int/bnd-api/vienna-classification-assistant/}

We apply a neural architecture search algorithm out of the box to learn multi-label classification using Vienna categories. Specifically, a 20-layer pretrained NASNET \citep{DBLP:journals/corr/ZophVSL17}, training half of its layers. We obtain an accuracy of 0.1155 in the level 3 of the hierarchy and an accuracy of 0.2786 in the level 2, showing the difficulty of the task.

%0.2786 accuracy level2
%0.1155 accuracy level3

%utilizando una NASNET pretrained training 
% https://github.com/asier-gutierrez/tm-dataset/blob/main/baselines/tm_multi_classification/nasnet.py#L17
% https://github.com/asier-gutierrez/tm-dataset/blob/main/baselines/tm_multi_classification/main.py#L35

\section{Conclusions \& Future Work}
The dataset presented in this paper is a continuously growing dataset of figurative images that is manually annotated.% annotated by humans in a very conscientious way.% Its usefulness lies in measuring the interpretation that Deep Learning systems have on images and texts with respect to humans.

This dataset can be used to generate systems that proceed in the artistic generation of logos in the same way as humans do, or to develop systems that understand the abstract artistic capacity of humans.

We provided simple baselines for two tasks. We expect that the scientific community will creatively contribute new ways to use this dataset either as a benchmark or to develop downstream applications.

As future work, we suggest \begin{enumerate*}
    \item extending the dataset with information from other offices (see WIPO), \item segmenting the images according to their figurative marks and textual elements to conform two separate datasets ("pure" images and logos), \item applying the dataset to other tasks (such as OCR), \item aligning Vienna classification with other existing classifications (e.g., CIFAR), and \item Outperforming the discriminative and generative baselines we provided.
\end{enumerate*}

\bibliographystyle{apalike}  

\clearpage
\bibliography{references}

\end{document}